\documentclass[journal]{IEEEtran}
\usepackage{wasysym}        % For clock symbol
\usepackage{fontawesome5}    % For gear and scales symbols
\usepackage{amssymb}         % For diamond symbol
\usepackage{pifont}          % For lightning symbol

\usepackage{subcaption} 
\ifCLASSINFOpdf
\else
\fi

\usepackage{graphicx}
\begin{document}
\title{Enhancing 5G O-RAN Communication Efficiency Through AI-Based Latency Forecasting}

\author{Raúl Parada,~Ebrahim Abu-Helalah,~%~\IEEEmembership{Member,~IEEE,}
        Jordi Serra,~\IEEEmembership{Senior Member,~IEEE,}~
        Anton Aguilar,~
        and~Paolo~Dini,~\IEEEmembership{Senior Member,~IEEE}% <-this % stops a space
        \vspace{-6mm}
\thanks{The authors are researcher at the SAI research unit within the Technological Telecommunications Centre of Catalonia (CTTC/CERCA)e-mail: \{rparada, aebrahim, jserra, aaguilar, pdini,\}@cttc.es.}}
\maketitle

\begin{abstract}
The increasing complexity and dynamic nature of 5G open radio access networks (O-RAN) pose significant challenges to maintaining low latency, high throughput, and resource efficiency. While existing methods leverage machine learning for latency prediction and resource management, they often lack real-world scalability and hardware validation. This paper addresses these limitations by presenting an artificial intelligence-driven latency forecasting system integrated into a functional O-RAN prototype. The system uses a bidirectional long short-term memory model to predict latency in real time within a scalable, open-source framework built with FlexRIC. Experimental results demonstrate the model's efficacy, achieving a loss metric below 0.04, thus validating its applicability in dynamic 5G environments. %Furthermore, we demonstrate how these latency predictions directly contribute to optimizing communication efficiency through real-time network adjustments and resource allocation.
\end{abstract}

\vspace{-2mm}

\IEEEpeerreviewmaketitle

%\vspace{-1.5mm}
\section{Introduction}
\vspace{-1.5mm}
\IEEEPARstart{O}{ptimizing} 5G open radio access networks (O-RAN) are critical to achieving low latency, high throughput, and reliable communication in dynamic wireless environments. Traditional methods, which are based on static models, do not adapt to these rapidly changing conditions. Recent advances in machine learning (ML) show promise in addressing these challenges, particularly for traffic forecasting and resource management. However, much of the existing research lacks real-world implementation and validation of scalability.
This work addresses this gap by introducing a latency forecasting system integrated into a fully functional O-RAN prototype. The system leverages artificial intelligence (AI)-driven xApps and the FlexRIC framework \cite{flexric} for real-time decision making, enabling programmable and scalable optimization. Our key contributions are as follows:
\begin{itemize} \setlength{\itemsep}{-0.3em} 
\item To design a bidirectional long short-term memory (LSTM) model for accurate latency prediction in dynamic 5G O-RANs.
\item To validate the system as a functional O-RAN prototype using open-source tools.
\item To develope a containerized framework for modular and reproducible O-RAN deployment.
\end{itemize} 

In existing research, ML techniques have been applied to optimize O-RAN, with a focus on traffic forecasting, latency reduction, and resource management. 
Habib et al. \cite{Habib2024} use LSTM models for traffic forecasting and dynamic scaling in 5G O-RAN, but lack hyperparameter optimization. Kavehmadavani et al. \cite{Kavehmadavani2023} employ time-series models like autoregressive integrated moving average (ARIMA) and LSTM for latency reduction, but do not evaluate model configurations or implement a hardware prototype. Khalid et al. \cite{Khalid2023} compare ML models for network optimization, but their work is limited to simulations. Khalid and Manar \cite{Khalid2024} use reinforcement learning for dynamic resource management, focusing on policy learning rather than real-time latency forecasting. Perveen et al. \cite{Perveen2023} address the integration of traffic prediction with network optimization but emphasize standard models over scalability techniques or latency trade-offs. In contrast, our work implements a real-time latency predictor for further optimal transmission decision, validated through a hardware-based O-RAN prototype. This approach combines scalability and practical deployment, addressing gaps in prior research \cite{Habib2024, Khalid2023}. Table \ref{tab:comparison} compares our work with the above-mentioned papers:
\begin{table}[h]
    \centering
    \renewcommand{\arraystretch}{0.8} % Adjust this value to change row height
    \caption{\small Comparison of our contributions to related works based on Latency Prediction (LP), Prototype Implementation (PI), Model Complexity (MC), Real-Time Processing (RTP), and Latency Trade-off (LTo).}
    \begin{tabular}{|l|c|c|c|c|c|}
        \hline
        \textbf{Work} & \textbf{LP} & \textbf{PI} & \textbf{MC} & \textbf{RTP} & \textbf{LTo} \\
        \hline
        \textbf{Habib et al.\cite{Habib2024}}         & $\checkmark$     & --               & $\rightarrow$    & --               & $\rightarrow$    \\
        \hline
        \textbf{Kavehmadavani et al.\cite{Kavehmadavani2023}} & $\checkmark$     & --               & $\rightarrow$    & $\rightarrow$     & $\rightarrow$    \\
        \hline
        \textbf{Khalid et al.\cite{Khalid2023}}       & $\checkmark$     & --               & $\rightarrow$    & --               & $\downarrow$     \\
        \hline
        \textbf{Khalid \& Manar\cite{Khalid2024}}     & --               & --               & $\uparrow$       & $\checkmark$     & $\uparrow$       \\
        \hline
        \textbf{Perveen et al.\cite{Perveen2023}}     & $\checkmark$     & --               & $\downarrow$     & $\rightarrow$     & $\rightarrow$    \\
        \hline
        \textbf{Our Work}              & $\checkmark$     & $\checkmark$    & $\uparrow$       & $\checkmark$     & $\uparrow$       \\
        \hline
    \end{tabular}
    \label{tab:comparison}
\end{table}\newline
This paper uniquely bridges the scalability and real-world validation gap by deploying a fully operational hardware prototype with AI-driven latency forecasting capabilities, demonstrating significant performance improvements over existing solutions.
\vspace{-4.5mm}
\section{Latency forecasting architecture}
\begin{figure}[!ht]
    \centering
    \begin{subfigure}[t]{\linewidth}
        \centering        \includegraphics[width=1.0\linewidth]{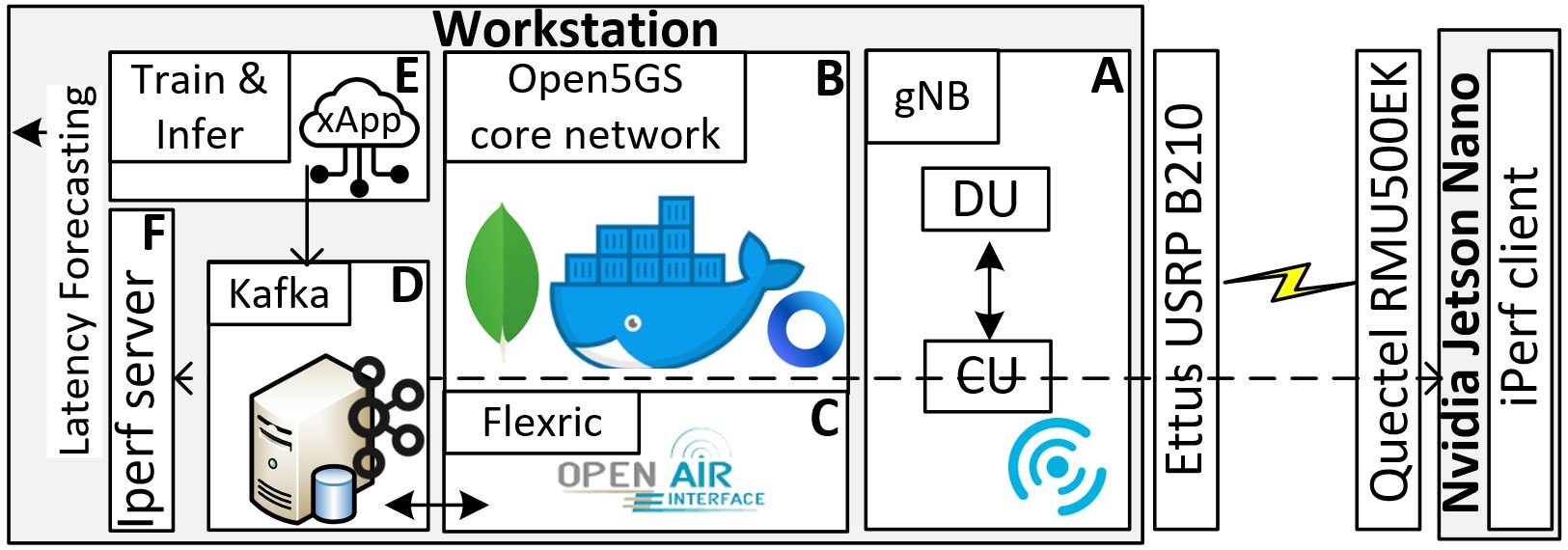} % Replace with the first image filename
    \end{subfigure}
    \vspace{-2mm} % Adjust spacing between the images
    \begin{subfigure}[t]{\linewidth}
        \centering        \includegraphics[width=1.0\linewidth]{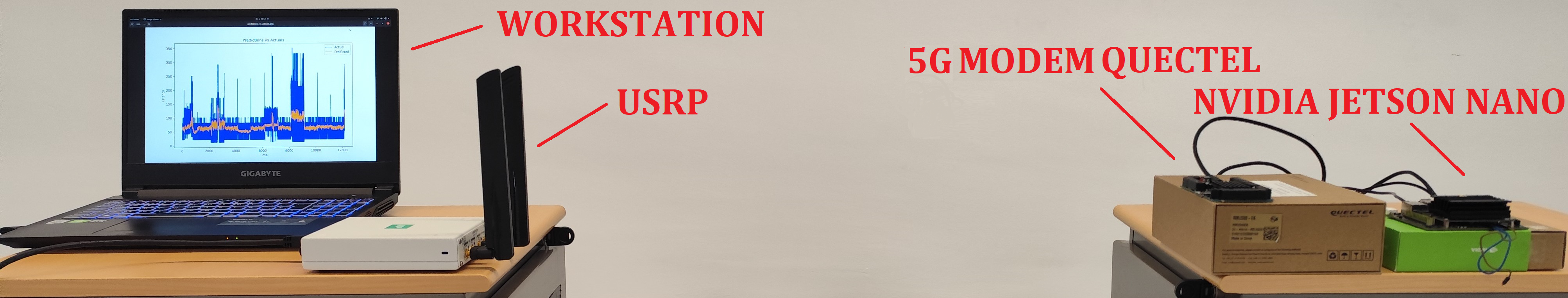} % Replace with the second image filename
    \end{subfigure}
    \caption{\small The top image illustrates the latency forecasting architecture and the bottom image shows the real demo.}
    \label{fig:arch}
\end{figure}

\vspace{-1.5mm}
This section describes the composition of the entire latency forecasting architecture. The top image in Figure \ref{fig:arch} illustrates the complete architecture. The hardware elements of this setup are: a GPU-based workstation (Debian 12), a software defined radio device (Ettus USRP B210), an Nvidia Jetson Nano (Ubuntu 18) and a 5G modem (Quectel RMU500EK). To enhance flexibility and scalability, we deployed Linux containers (LXCs) on the workstations. All the tools used are open source, ensuring reproducibility. The LXCs are: \textbf{A)} srsRAN Project (Ubuntu 22): The O-RAN-native centralized unit (CU) / distributed unit (DU) developed by SRS that acts as a stand-alone gNB, \textbf{B)} Open5GS (Ubuntu 22): A 4G/5G core network implementation, \textbf{C)} Flexric (Ubuntu 24): A RAN intelligent control framework for programmable, real-time control and optimization of 5G RANs, \textbf{D)} Kafka (Ubuntu 24): A consumer instance that stores the key performance metrics (KPMs) generated by Flexric, \textbf{E)} xApp (Ubuntu 20): An edge service instance that trains the selected KPMs from Kafka consumer and infers latency forecast measurements and \textbf{F)} iPerf server (Ubuntu 20): A network testing tool that listens for incoming iPerf connections from iPerf clients to generate traffic and provide valid KPM values. 
\vspace{-2.5mm} 
\section{Experimental setup \& results} \label{sec:res}
\vspace{-2mm}
This section provides the steps on how to set up the latency forecasting architecture described in the previous section and some preliminary experimental results. The bottom image from Figure \ref{fig:arch} shows the demonstration setup in our facilities' laboratory. The demo runs as follows (assuming you have created all the LXCs mentioned above): Execute \textit{lxc list} to display the created LXCs, their state should appear stopped. First, we launch containers A, B, C and D to run the O-RAN environment with \textit{lxc start gnb o5gs ric kafka}. The 5G Core instance might start automatically in the background. This is done to reduce the number of terminals to check. We enter the gNB container and execute \texttt{./run} to start the base station. Then, inside the Jetson Nano, we start the user equipment (UE) instance to make the O-RAN environment run. Once the O-RAN environment is running, we can start both the Kafka and FlexRIC instances to prepare the environment for KPM acquisition. Afterward, we start the xApp script, which allows training the acquired KPMs for a further inference period from the trained model. Since the xApp requests data from the Kafka consumer instance, we need to start it to connect to FlexRIC. Finally, because the system requires traffic flow throughout the system, we start the iPerf server waiting for an iPerf client to execute on the Nvidia Jetson Nano. The xApp forecasts latency in real time and displays a plot comparing actual and predicted values.
The system tracks several FlexRIC KPMs to evaluate and optimize network performance. Key metrics include the number of connected UE devices to assess network load, past latency values for real-time performance, and physical resource blocks (available and total) uplink for resource capacity. Transmission reliability is monitored via uplink packet success rate, while uplink throughput is reflected in transmitted service data unit volumes and throughput metrics. Air interface delay captures transmission latency, and signal quality is measured by signal-to-noise and channel quality indicator, enabling adaptive optimizations such as modulation and coding adjustments.
Since the data are streamed in time-series format, we have chosen the LSTM algorithm to train a prediction model to infer the latency forecasting, we have used the python library TensorFlow due to its hardware adaptability for process performance optimization. The training is performed offline with a preliminary dataset composed of 56k rows. The model is configured as bidirectional LSTM with 100 units and ReLU activation, leveraging a 60-step lookback period. The model includes dropout (0.2) to prevent overfitting and outputs a single value using a dense layer. It is compiled with mean squared error loss and the Adam optimizer (learning rate = 1e-5). Training employs early stopping (patience = 10) and saves the best model, using validation loss as the metric. Both training and validation loss metric achieved were below 0.04 indicating that the model is not overfitting, and it generalizes well to new, unseen data. The latency prediction contributes to a more efficient communication by deciding whether to transmit or not based on the link quality, hence, reducing the number of possible unsuccessful transmissions. A real-time prediction can be observed and described in the video uploaded in https://gitlab.cttc.es/supercom/LatencyForecasting.
\vspace{-2.5mm}
\section{Conclusion}
This work demonstrates the feasibility and effectiveness of integrating AI-driven latency forecasting into 5G O-RAN using open-source tools and hardware prototypes. Using a bidirectional LSTM model for real-time prediction, the system optimizes network performance through dynamic resource management and adaptive decision making. The results achieved, including accurate latency prediction and scalable deployment, address critical gaps in existing approaches. This implementation paves the way for advanced O-RAN optimizations, improving the reliability and efficiency of next-generation wireless networks. Future work includes exploring state-of-the-art models such as extended LSTM and temporal Kolmogorov-Arnold networks, measuring the system's CO2 emission reduction, and scaling the demo with multiple UEs by uniquely connecting additional 5G modems.
\vspace{-3.5mm}
\section*{Acknowledgment}
This work has been funded by the ”Ministerio de Asuntos Económicos y Transformación Digital” and the European Union-NextGenerationEU in the frameworks of the ”Plan de Recuperación, Transformación y Resiliencia” and of the ”Mecanismo de Recuperación y Resiliencia” under references TSI-063000-2021-18/24/77. The authors thank Marco Oliveira for designing the original demo in the SAI research unit at the CTTC.

\ifCLASSOPTIONcaptionsoff
  \newpage
\fi



\vspace{-4mm}
\bibliographystyle{IEEEtran}
\bibliography{bibtex/bib/IEEEexample}
\end{document}